\documentclass[letterpaper, 10 pt, conference]{ieeeconf}  %

\IEEEoverridecommandlockouts     
\usepackage{graphicx}
\usepackage{amsmath,amssymb,amsfonts}
\usepackage{caption,subcaption}
\usepackage{url}
\usepackage{booktabs}
\usepackage{cite}

\graphicspath{ {images/} }

\title{\LARGE \bf Experimental Evaluation of a Pseudo-Doppler Direction-Finding System for Localizing Radio Tags}

\author{William E. Gerhard III and Pratap Tokekar%
\thanks{W. Gerhard III was with the Department of Electrical
and Computer Engineering, Virginia Tech, Blacksburg, VA 24061 USA when part of the work was completed.  (email: {\tt\small willig42@vt.edu}).}
\thanks{P. Tokekar was with the Department of Electrical
and Computer Engineering, Virginia Tech, Blacksburg, VA 24061 USA when part of the work was completed. He is currently with the Department of
Computer Science, University of Maryland, College Park, MD 20742, USA (email: {\tt\small tokekar@umd.edu}).}}

\begin{document}
\maketitle
\thispagestyle{empty}
\pagestyle{empty}

\begin{abstract}
We present the design of a radio antenna system for obtaining instantaneous bearing measurements towards a radio emitter. Our work is motivated by applications where robots are used for localizing and tracking radio-tagged wildlife. The traditional method is to use directional antennas that need to be rotated in order find the bearing which is time consuming. Instead, we present a low-cost system capable of finding bearing measurements almost instantaneously using an antenna array. This is particularly appealing for wildlife tracking with Unmanned Aerial Systems (UASs) where remaining stationary can be challenging and energy consuming, in addition to being slow. The proposed system uses existing open source hardware and software systems  and leverages principles of pseudo Doppler direction-finding. The resulting system was tested in an anechoic chamber and in outdoor settings. The outdoor tests with particle filtering show that the resulting system is capable of localizing radio tags within 5 meter accuracy starting with an initial estimate of 200m x 200m. 
\end{abstract}

\section{Introduction} \label{ch:introduction}
	
Unmanned Aerial Systems (UAS) have many applications in field biology, and their potential as a tool for wildlife tracking in rugged, difficult to access terrain has been demonstrated in several studies. In particular, \cite{linchant_2015} provides a comprehensive survey of how UASs are changing the monitoring and tracking of populations of wildlife. 
Beyond simply replacing manned aircraft, UASs are a potent tool for the active tracking and localization of wildlife. The ability to monitor where wildlife go, how they move and react to changes in the environment and other behaviors not normally observed are critical to conservation efforts and to the management of wildlife populations for future generations. 


Several recent works have explored using the unique capabilities offered by UAS combined with radio direction-finding techniques to track wildlife.  Vonehr et al.~\cite{vonehr_2016} utilized a rotating multirotor UAS with a directional antenna to determine the angle of incidence of an incoming signal. 
Bayram et al.~\cite{bayram_2017} expanded upon the idea of using directional antennas on-board multirotor UAS by creating a path planning algorithm that enabled three UAS to work together to triangulate the signal source and calculated its position. Their algorithm  was able to estimate the signal’s source with positional errors of 3.7 meters, 10.7 meters, and 18.2 meters, depending on the starting parameters and the experiment setup. More recently~\cite{bayram_2018}, the same group explored the use of a single UAS with a directional antenna to estimate the position of a signal source. By using an online strategy to choose measurement locations, the UAS was able to localize the signal source to a user-specified 10 meters of error. Previously, a similar setup (rotating antenna to find bearing) was used on a robotic boat to localize radio-tagged fish~\cite{isler2015finding,vander2012cautious,tokekar2013tracking}. The disadvantage of this method is that it takes in the order of minutes to find a single bearing measurement since the antenna has to sweep through multiple angles to find the direction with the highest signal strength. 

An alternative is to directly use the received signal strength and extrapolate the signal's origin from its gradient. Bayram et al.~\cite{bayram_2016} developed a local search algorithm that drives the UAS towards the signal source through measuring characteristics of the received signal. By using this algorithm in an open field, the signal source was able to be located with a positional error of 50 meters. Other work in this area done by \cite{desrochers_2018} produced similar results with tests conducted in boreal forests of Canada, but with a larger error of 94 meters. 


\begin{figure}[htb!]
						\centering
						\includegraphics[height=1.25in]{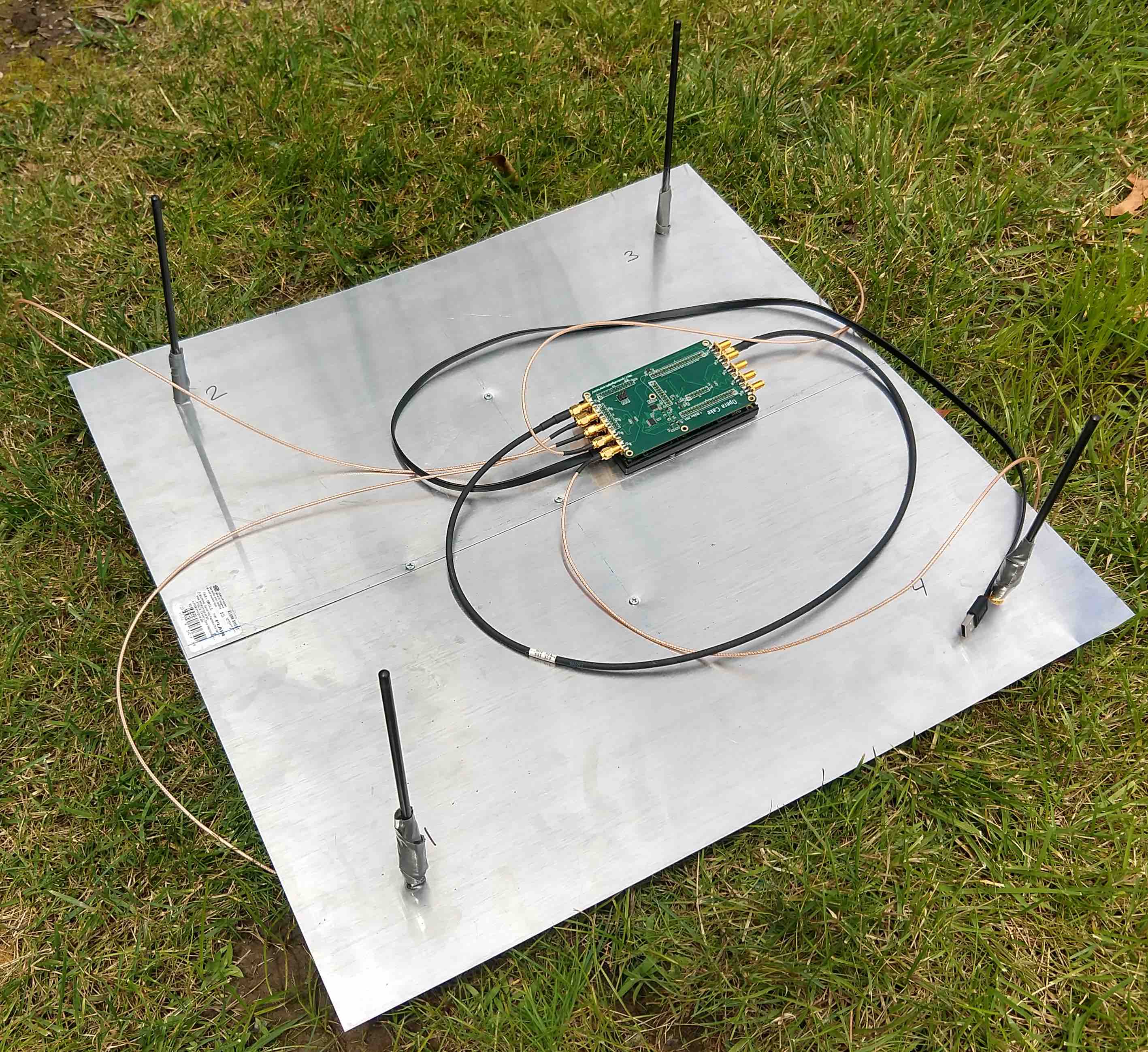}
						\includegraphics[height=1.25in]{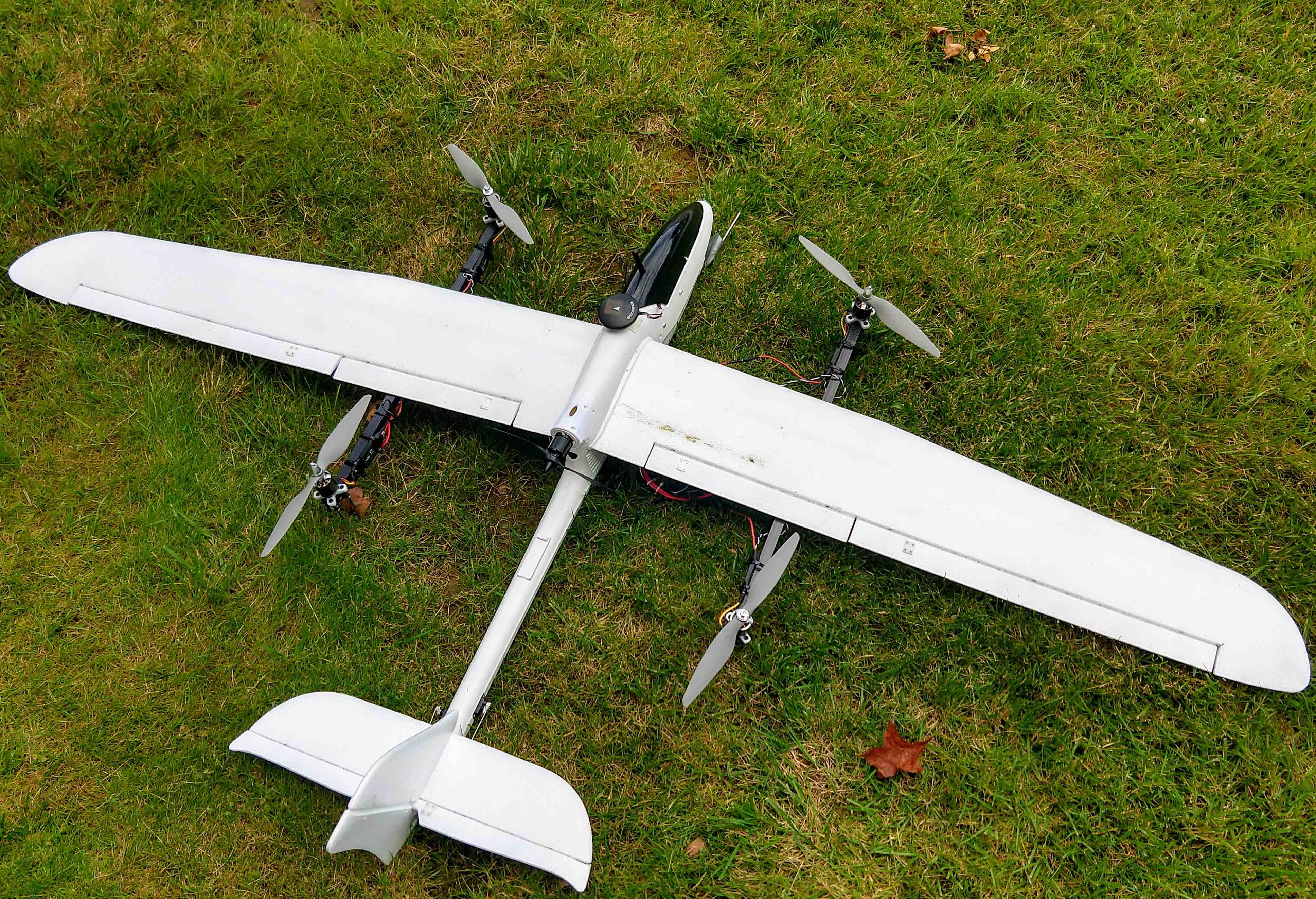}
						\caption{The PDDF system with four antennas and a HackOne RF Software Defined Radio. The four antennas (without the base plate) can be mounted on the arms of the hybrid UAS developed.} 
						\label{fig:completepddf}
					\end{figure} 

Bearing measurements are more informative than the gradient-based methods, especially in the presence of multi-path effects. To mitigate the limitation of slow measurements, in this paper we leverage a design with four antennas and use the relative signal between them to extract the bearing. This design is motivated by applications where the antennas are to be mounted on a UAS. We employ a method termed as Pseudo-Doppler Direction-Finding (PDDF) that uses the four antennas along with a Software Defined Radio (SDR) to compute the measurement (Figure~\ref{fig:completepddf}). The resulting system was extensively tested in the lab and the field.

Also, through the development and use of the PDDF, this work enables orders of magnitude more angle of incidence data to be collected than any other method discussed, which speeds upsampling. Furthermore, without a need to hover in place, the system can be mounted on a fixed-wing UAS instead of a multirotor UAS. This improves estimator performance and system flight range and duration. We are using a hybrid UAS (Figure~\ref{fig:completepddf}) which can take-off and land vertically but can fly using the fixed-wing mode for increased endurance.

To summarize, this work contributes to the field through:
\begin{itemize}
\item Development of a low-cost software defined radio doppler direction finder using Commercial, Off-The-Shelf (COTS) products; and
\item Validation that the resulting system along with a particle filter fusing bearing measurements can localize a radio transmitter within 5m error (in an area of 200m x 200m). 
\end{itemize}
The potential of PDDF for wildlife tracking was mentioned by MacCurdy et al.~\cite{maccurdy2011automated}. However, no system design or experimental results were provided. This paper contributes to the wildlife tracking literature by presenting a PDDF system suitable for UAS applications, integrating it with particle filter for tracking wildlife and presenting an experimental evaluation of the system. All software and supporting documents can be found at: \url{https://github.com/raaslab/PDDFThesis}

\section{Background on Pseudo-Doppler Direction-Finding} \label{ss:pddfB}

Doppler direction-finding utilizes the Doppler effect to determine the direction of incidence. If an antenna element is rotated in a circle of radius \(R\), then the received signal \(\omega_0\) is frequency modulated with the rotational frequency \(\omega_r\) due to the Doppler effect. As the antenna rotates towards the  direction of incidence, the received frequency increases; while when the antenna rotates away from the direction of incidence, the received frequency decreases~\cite{dopplerdf}.

					
This means the instantaneous amplitude of the received signal \(u(t)\) is related to both the antenna radius \(R\), direction of incidence \(\alpha\) and wavelength of the frequency of interest \(\lambda\). \(\theta\) is the phase content of the signal of interest~\cite{Rohde_Schwarz},
\begin{equation}
\label{pddfequation}
u(t) =arccos(\varphi(t)) =  arccos(\omega_0t+\dfrac{2 \pi R}\lambda cos(\omega_rt-\alpha)+	\theta)
\end{equation}

By taking the derivative of the instantaneous  \(\varphi(t)\), the Doppler signal can be extracted~\cite{Rohde_Schwarz},
\begin{equation}
\omega(t) = \dfrac{d\varphi(t)}{dt} = \omega_0-\dfrac{2 \pi R}\lambda sin(\omega_rt-\alpha).
\end{equation}

After removing the DC content \(\omega_0\), the demodulated signal \(S_D\) is obtained~\cite{Rohde_Schwarz}
\begin{equation}
\label{pddfequationsd}
S_D = \dfrac{2 \pi R}\lambda sin(\omega_rt-\alpha).
\end{equation}
Now, if the phase of the modulated signal \(S_D\) is compared to a reference signal \(S_r\) of equal center frequency to the antenna rotation frequency \(\omega_r\) then the angle of incidence \(\alpha\) can be extracted~\cite{Rohde_Schwarz},
\begin{equation}
S_r = -sin(\omega_rt).
\end{equation}

The Doppler direction finder still suffers from the same issues as the directional antennas --- mechanical rotation. Pseudo-Doppler direction-finding is an alternative approach to Doppler direction-finding that uses the same principles but replaces a single rotating antenna with an electronically sampled circular antenna array.  Each antenna is sampled in sequence at a fixed frequency, equivalent to \(\omega_r\). By combining the samples into a single signal, the resulting superimposed phase shift is the same as a single, physically rotating antenna. This means the angle of incidence can still be determined by comparing the phase of the demodulated Doppler signal to a fixed frequency sinusoid. \cite{dopplerdf}

The advantages of Pseudo-Doppler direction-finding is threefold. By keeping the antenna spacing under \(\lambda/2\) (or equivalently keeping radius R under \(\lambda/4\)) the angle of incidence can be determined unambiguously, and an antenna array of any size is possible. This allows the Pseudo-Doppler direction-finding system to have high immunity to multi-path reception and relatively high sensitivity. In addition, this method only requires a single RF processing channel and a method of antenna switching to operate. \cite{Rohde_Schwarz}

	\section{Design of the Pseudo-Doppler Direction Finder} \label{s:pddfsd}

While few professional and hobbyist grade COTS PDDF systems exist, all of them use an external Very High Frequency (VHF) radio receiver, which limits the frequency flexibility and software applications. No existing COTS system for pseudo-doppler direction-finding existed when this work was started. However, the components to build such a system did exist. 
	

At a minimum, to track tagged wildlife the PDDF had to operate in the 145 to 165 MHz bands. This VHF band is commonly used by radio collars designed for larger wildlife, while the Ultra High Frequency (UHF) band is more typically used to track smaller wildlife such as birds due to reduced antenna size. Therefore the PDDF should be flexible and re-configurable for tracking different targets in different radio bands. In addition, the system needs to be small, low power and light to maximize the UAS’s flight time and fit within its size requirements. Open-source software was required to increase accessibility to the system and enable other interested parties to recreate the work.

\subsection{Selection and construction} \label{ss:scsdr}

Our PDDF system design is based around the COTS HackRF One~\cite{HackRFOn58:online} Software Defined Radio (SDR) and the Operacake add-on board~\cite{OperaCak4:online}. The HackRF One has a 1 MHz to 6 GHz operating frequency with a half-duplex transceiver and takes up to 20 million samples per second. 

\begin{figure}[t]
\centering
\includegraphics[width=0.5\columnwidth]{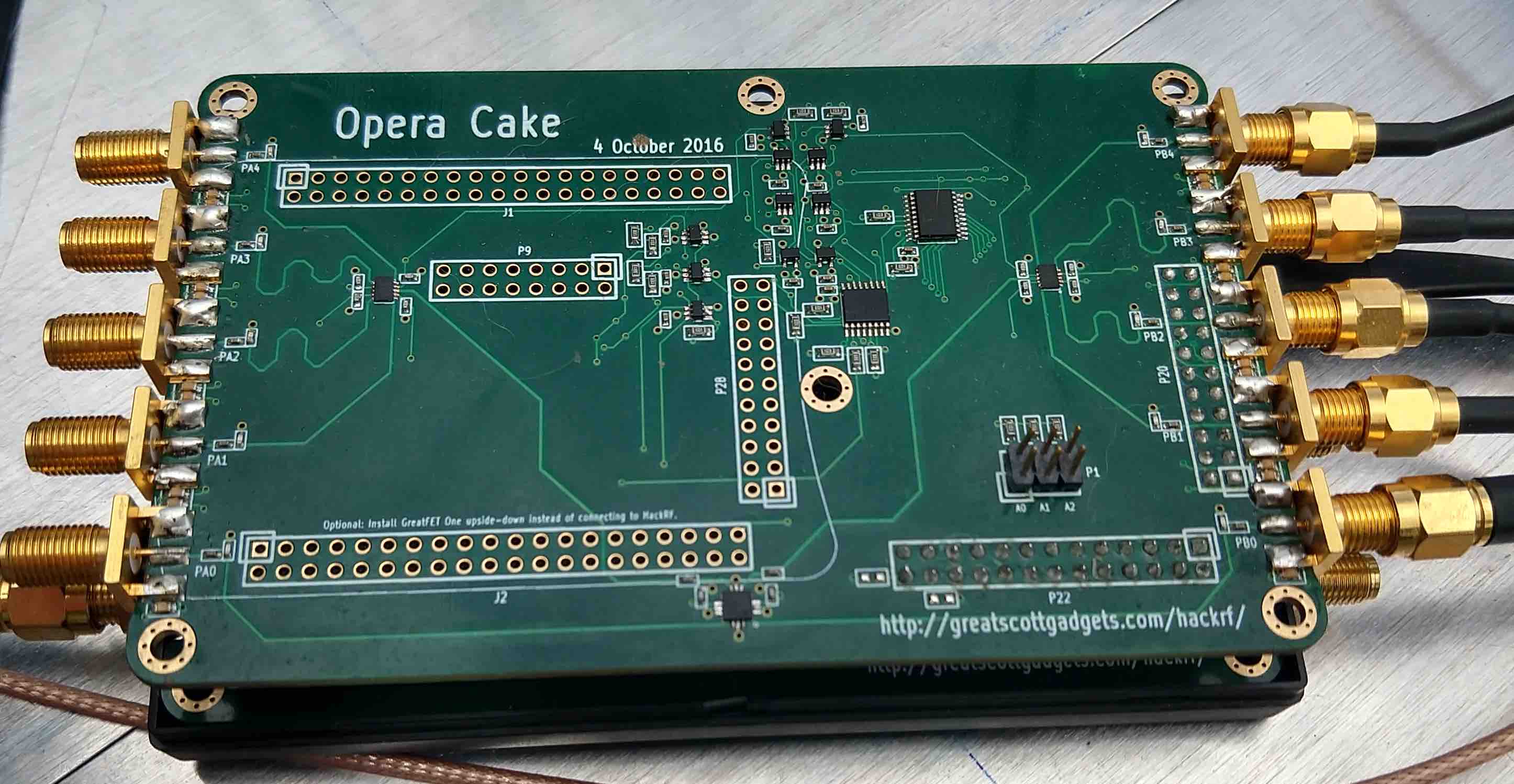}
\caption{Assembled Operacake board} 
\label{fig:operacake}
\end{figure} 
					
In addition, the Operacake add-on board enables the HackRF One to switch between one of eight RF inputs. Due to how the Operacake is interfaced with the HackRF One, the switching between antennas is transparent to the onboard ADC and happens in between samplings.  Firmware modifications to the HackRF were required to ensure proper operation of the Operacake. The result was an automatic start of the antenna switching, with a switching rate tied to the data sample rate \(S_{hackRF}\) and the number of samples per antenna \(Samples_{antenna}\): 
\begin{equation}
\label{eq:rotationfreqcal}
F_{AntennaSwitching} = \dfrac{2S_{hackRF}}{4Samples_{antenna}}.
\end{equation} 

The final component of the PDDF system was the antenna array. In order to fulfill the spatial Nyquist theorem, the antenna array cannot exceed \(\lambda/2\) spacing between each antenna element, i.e., a square four element antenna array cannot exceed \(0.35\lambda\) a side. However, experiments conducted over the last couple of decades by the amateur radio fox hunting community have shown a square antenna array with sides not more than \(0.22\lambda\) provides optimal results for pseudo Doppler direction-finding~\cite{moell_2008}. The primary frequency of interest in this work is 150MHz, resulting in a \(\lambda\) of 78.72 inches (1.9986 meters) and therefore results in an antenna spacing of 17.3184 inches (0.44 meters). This spacing covers frequencies from 120 MHz to 170 MHz, which contains all the VHF frequencies used for wildlife tracking.

	\subsection{ GNURadio Flow Graph} \label{ss:gnuradiograph}
	
\begin{figure}[htb!]
						\centering
						\includegraphics[width=.95\columnwidth]{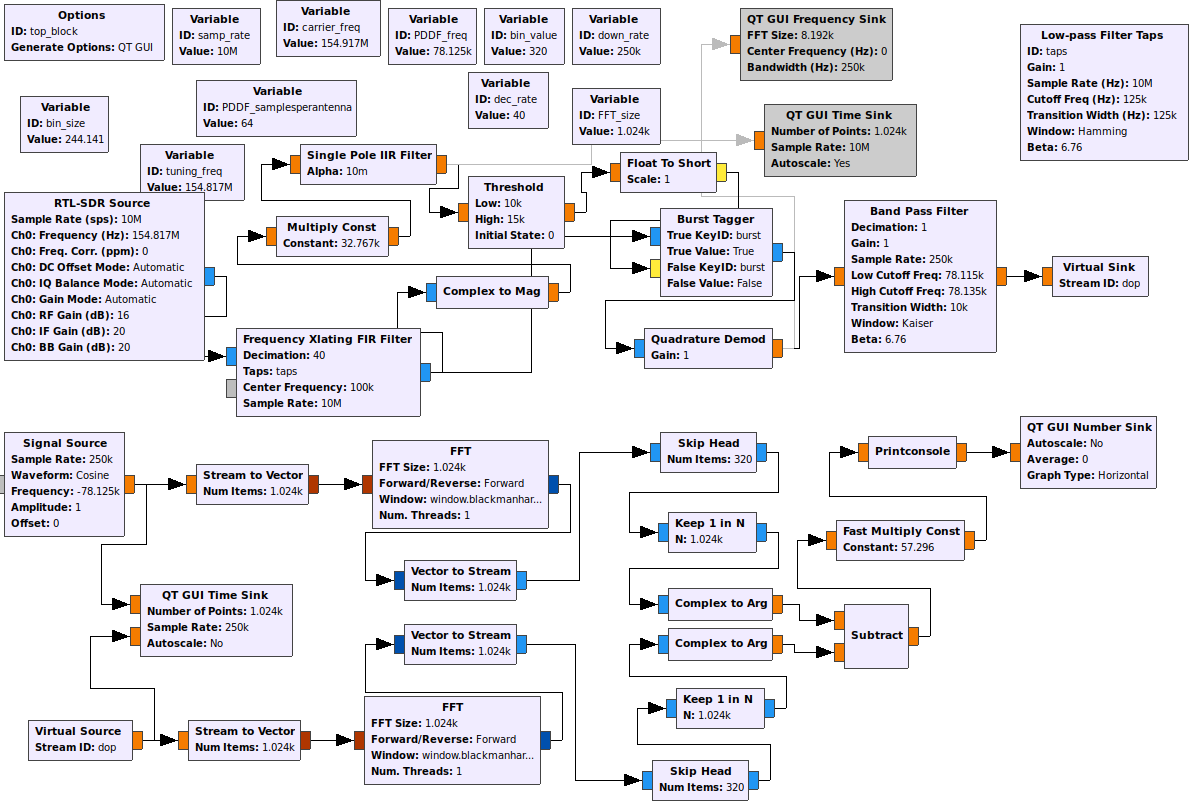}
						\caption{GNU Radio Flowgraph} 
						\label{fig:gnuradioflow}
					\end{figure}

	
	GNURadio was used to develop the pseudo-doppler direction-finding software (Figure~\ref{fig:gnuradioflow}), which is composed of four main sections: input, filtering, FFT calculation, and comparison.
	
The input section of the flowgraph consists of two blocks, a read block and a frequency translating FIR filter block. The FIR filtering block is required as the HackRF does not have a DC blocking filter built into the hardware. By using a frequency translating FIR filter, the HackRF One can be tuned to a frequency close to the actual frequency desired, and the signal can then be filtered and shifted in software, eliminating the DC offset. The filter also has built in decimation, allowing the HackRF One to be set to a higher sample rate to achieve the Operacake switch  frequency desired, and then have the sample rate reduced to something more manageable for lower performance systems. 

The filtering section of the flowgraph is a simple notch filter. The notch filter width is set to a tenth of the FFT bin size around the Operacake rotation frequency. This is intended to help suppress adjacent frequencies' random noise on the calculated phase.

To determine the angle of incidence of the incoming signal, the phase of the now filtered sinusoid at the Operacake rotation frequency must be compared to the phase of a non-offset cosine at the same frequency. An FFT is used to do this here where \(N\) is the length of the array \(x_k\), \(n\) is the current index in the array \(x_k\) and \(k\) is the sample rate in Hz:
\begin{equation}
X_k = \sum_{n=0}^{N-1} x_k. e^{\dfrac{-i2\pi}N kn} 
\end{equation}


Based on parameters set in the flow graph, the signals of interest fall exactly into the 320 FFT bin. The imaginary component \(\theta\) of the bin is extracted and compared and the difference between the two values is the angle of incidence \(\alpha\) in radians, $\alpha = \theta_1-\theta_2$. Finally, a custom GNURadio block logs the calculated angle of incidence as a number along with a microsecond UNIX timestamp, and saves it in a CSV file for later use.



\subsection{Particle Filter} \label{s:particlefilter}
	
While the PDDF system provides angle of incidence information, that information by itself is unable to localize the transmitter. To determine the location of the transmitter, additional angle of incidence information is needed at the same time interval from different spatial locations as demonstrated by~\cite{bayram_2017}, or multiple angles of incidence measurements at different spatial and temporal locations, with the strong assumption that the target did not move, as done in more recent work by~\cite{bayram_2018}. A third approach using particle filters, that does not require the strong assumption that the target does not move, is used in this work. By combining multiple angle of incidence measurements at different spatial and temporal locations with models representing sensor and target behaviors, an estimated position and confidence in that estimate can be calculated. 


In out particle filter implementation, the state space \(X_n\) is five dimensional: 2D  position, 2D velocity, and orientation relative to a global origin.
	
 
 

Due to the various sizes, habitats and habits of wildlife, developing motion models for how wildlife moves in a difficult task and a lot of research effort has focused on modeling how animals over~\cite{maarell2002foraging,benhamou2007many,bartumeus2005animal}. Since the particle filter is working on a short time scale, minutes instead of hours, the target is assumed to be moving in a random direction at a random speed at each time step, resulting a Brownian-like motion model. Thus:	
 \begin{align}
S_n&=S_{n-1}+\int_0^\Delta U_n+Z_n\\
U_{nx}
&=U_{n-1}^x+-\int_0^\Delta A_x+Q_n\\
U_{ny}
&=U_{n-1}^y+-\int_0^\Delta A_y+Q_n.
	 \end{align}
$U$ is the velocity, $S$ is the position, $Z_n$ is a random Gaussian distribution with a mean of 0 and a variance of .5, and $Q_n$ is a random Gaussian distribution with a mean of 0 and a variance of 25. $A$ is the measured acceleration on the PDDF system in the $X$ and $Y$ directions. This model makes the PDDF system’s current position the center of the particle filter's coordinate system, allowing a direct comparison between each particle's bearing to $(0,0)$ and the incoming corrected bearing measurement from the PDDF system. 

\begin{figure}[htb!]
						\centering
						\includegraphics[width=0.7\columnwidth]{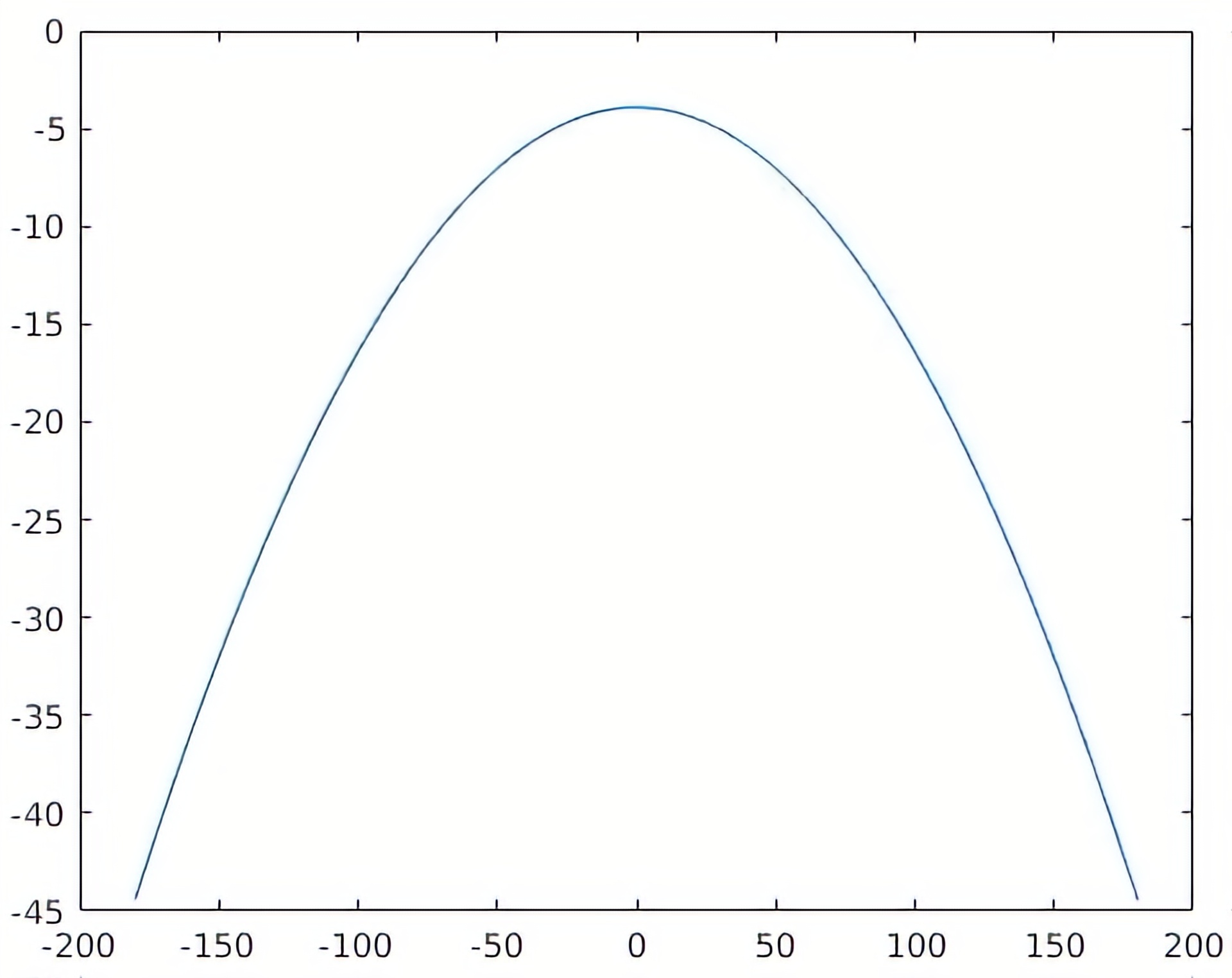}
						\caption{Plot of Sensor Model} 
						\label{fig:sensormodel}
					\end{figure}

	
The measurement model (Figure~\ref{fig:sensormodel}) is one dimensional and simply applies a likelihood estimate based on how close each particle's saved bearing is to the measured bearing. This results in:
 \begin{equation}
-0.5\log(2 \pi \sigma^2) - \dfrac{\theta_n^2}{2 \sigma^2} 	
 \end{equation}
where \(\sigma^2\) is the signal variance and  \(\theta_n\) is the difference between the particle's saved bearing and the measured bearing. This shape increases the likeliness of particles whose saved bearings are close to the measured bearing, while decimating particles whose bearings are 180 degrees off. The variance can be adjusted depending on real world data collection results to account for errors and other noise. 

	
The particle filter utilizes a sequential Monte-Carlo framework designed for robust, cross-platform development~\cite{johansen_2009}. The particle filter uses each incoming bearing measure to add or subtract from the likelihood of each particle being the correct location. and is re-sampled after 50\% of the particles have been eliminated~\cite{chen_1998}.
To get the estimated position of the transmitting signal, a simple approximation of the particle cloud's integral was used to find the mean and variance of \(s_n^x\)  and \(s_n^y\). 

\section{Results} \label{ch:results}
	
Each component designed and built for this work was tested individually to ensure proper operation. Outdoor testing of the PDDF system was conducted, and finally, the results were fed into the particle filter.

	\subsection{Anechoic Chamber Testing} \label{s:unittests}

\begin{figure}[htb!]
						\centering
						\includegraphics[width=0.7\columnwidth]{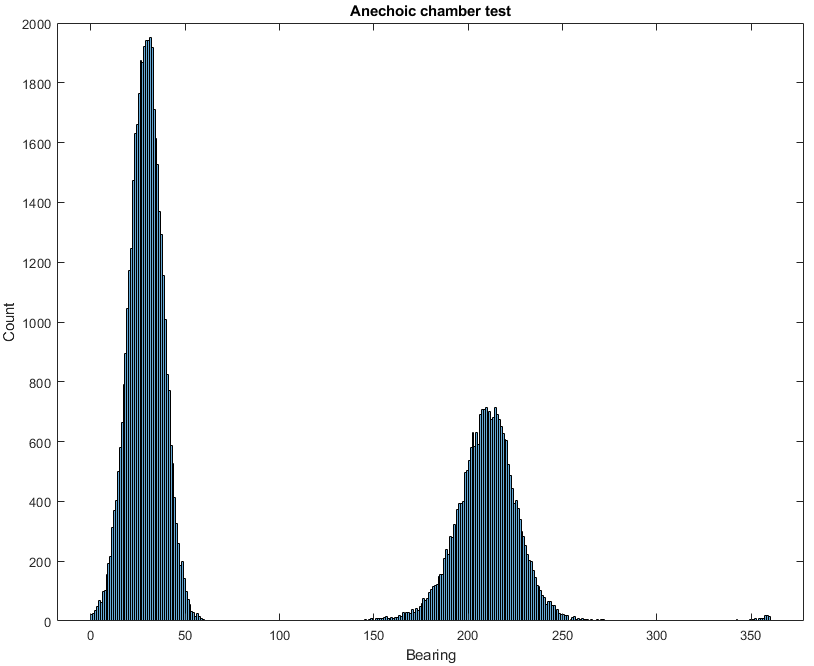}
						\caption{Histogram of angle of incidence calculations from anechoic chamber test}
						\label{fig:histochamber}
					\end{figure}

The PDDF system was tested using the anechoic chamber at Virginia Tech. A small antenna array was built using standard WiFi antennas spaced in a 1.2 inch (3 cm) radius circle. This array was connected to the Operacake and HackRF One and placed in the chamber. A -10 dB 2.4 GHz far-field signal was generated. The array was placed with antenna 2 approximately facing the far-field source and data collection commenced. After a few seconds, data collection was halted and the array was rotated approximately 180 degrees. The far field signal was generated and data collect commenced again. The resulting bearing estimates were plotted in a histogram. Figure~\ref{fig:histochamber} shows the 180-degree phase change, with two distinct signals close to 50 and 220 degrees. In addition the data showcased a Gaussian distribution of bearing estimates. 

\subsection{Moving PDDF Array Tests} \label{s:pddfparticletest}
\begin{figure}[htb!]
						\centering
						\includegraphics[width=.8\columnwidth]{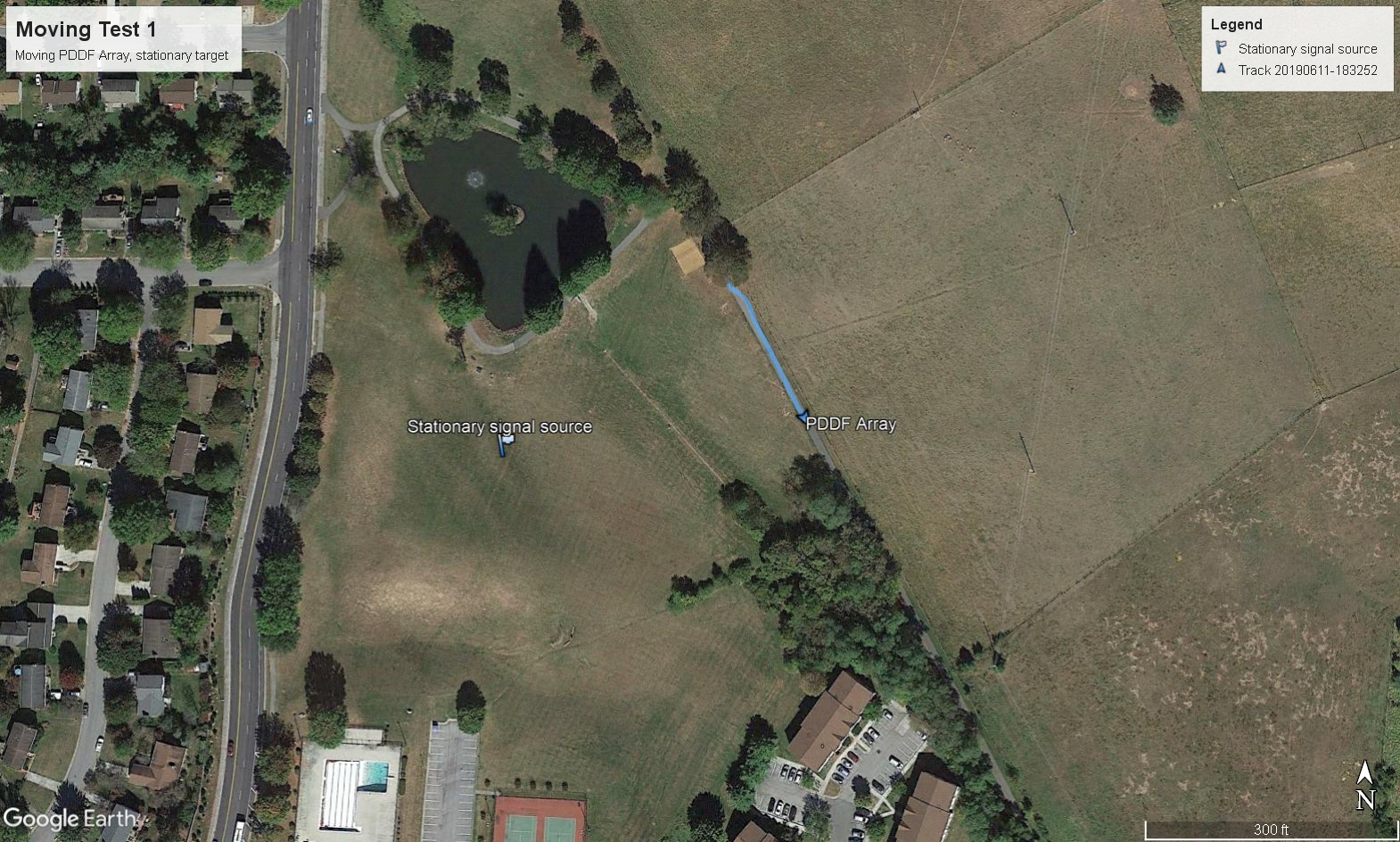}
						\caption{ Map showing the signal source and path of the PDDF system for the first particle filter data collection}
						\label{fig:particletest1}
					\end{figure}

We conducted a series of tests  to generate additional data to be fed into the particle filter. GPS was used to record the speed of the cart with the system along with its heading and position. For the first test, the PDDF array was placed on a cart which was pulled at a constant speed along a walking path. The signal source was stationary and located 100 meters away as shown in Figure \ref{fig:particletest1}.

\begin{figure}[htb!]
						\centering
						\begin{subfigure}[h]{0.95\columnwidth}
							\centering
							\includegraphics[width=.95\columnwidth]{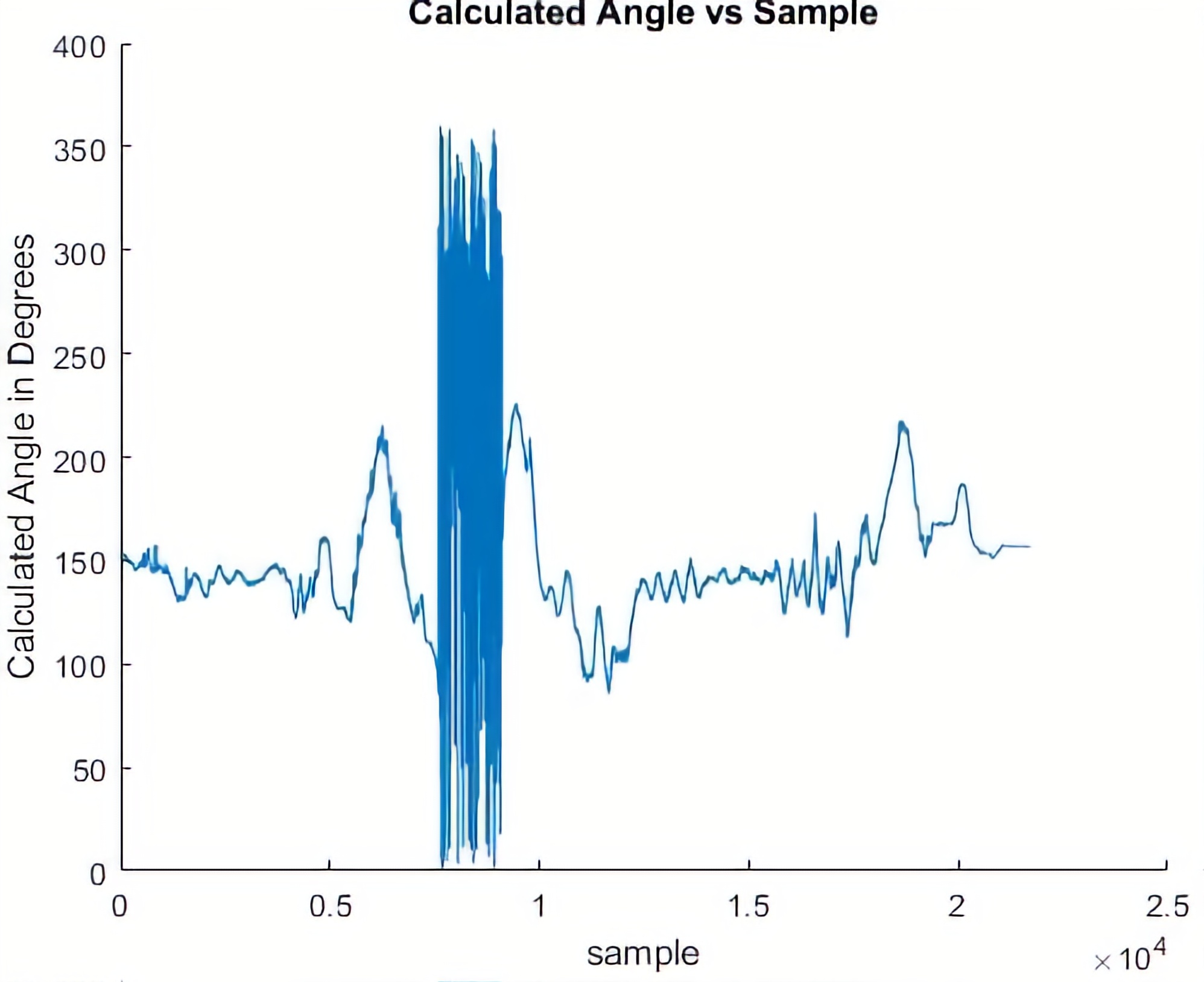}
							\caption{Calculated angle of incidence at full data rate}
							\label{sfig:particletest1movingfulldata}
						\end{subfigure}
						\begin{subfigure}[h]{0.95\columnwidth}
							\centering
							\includegraphics[width=.95\columnwidth]{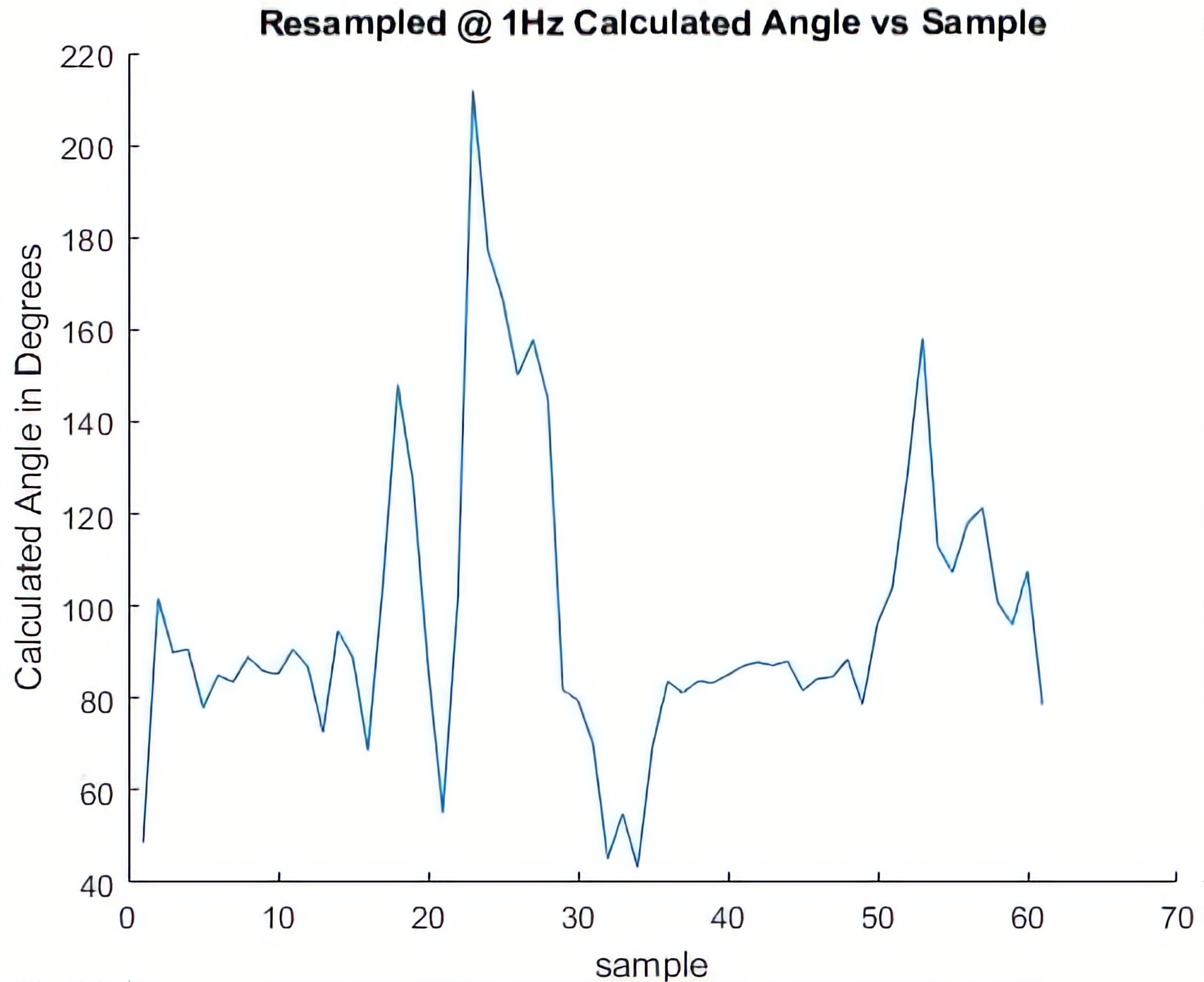}
							\caption{Calculated angle of incidence reduced to 1Hz}
							\label{sfig:particletest1moving1hx}
						\end{subfigure}
					\caption{Moving PDDF array data set with noise introduced}
						\label{fig:main_figuresubsampledparticledata1}
					\end{figure}

A third of the way into this test the broadcasting radio signal was turned off to introduce noise into the particle filter to see if it would be able to recover. This noise is visible in Figure \ref{sfig:particletest1movingfulldata} between 8,000 and 9,000 samples. The base data rate of the PDDF is approximately 330 Hz, which was down-sampled to 1Hz to match the GPS data. This down-sampled signal is shown in Figure \ref{sfig:particletest1moving1hx}. 

\begin{figure}[]
						\centering
						\includegraphics[width=.8\columnwidth]{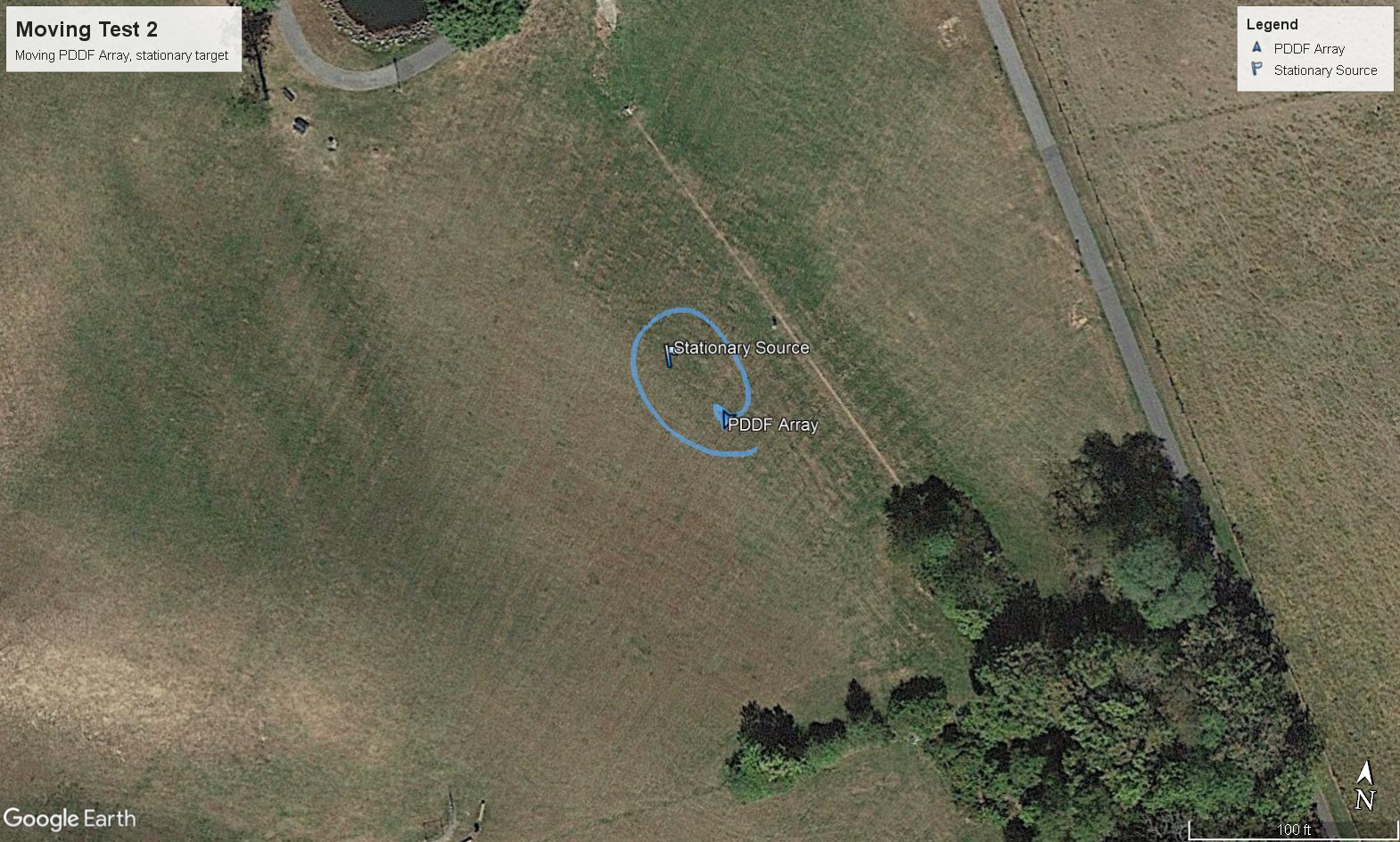}
						\caption{ Map showing the signal source and path of the PDDF system for test 2}
						\label{fig:particletest2}
					\end{figure}

This particle filter was run twice, once with a starting grid of 200m by 200m and once with a starting grid of 2km by 2km. The results were very close to identical and for brevity, only the 2km by 2km results are shown below.
							\begin{figure}
						\centering
					
		\includegraphics[width=.8\columnwidth]{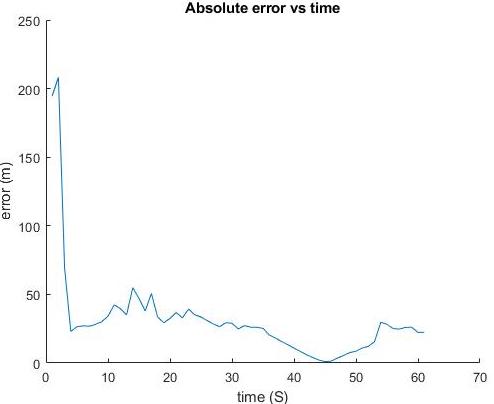}
					\caption{Particle filter results of data set 1 on a 2000 meter by 2000 meter grid. }
						\label{fig:main_figureparticlefilter1results2000}
					\end{figure}
Even with the noise added to the data set, the particle filter was able to localize the transmitter within 40 meters for the entire test, as shown in Figure \ref{fig:main_figureparticlefilter1results2000}. The particle cloud itself started with a uniform distribution and was able to converge. There is an offset between the true position and the estimated position in part due to the noise added to the system intentionally and from the antenna geometry changing from both the motion of the cart and vibrations from rolling over obstacles. Geometric dilution of precision also played a part, exaggerating angle errors the closer the PDDF got to the target. The filter exhibits the same tendency as previous tests to overestimate the distance to the target in the X dimension. The best way to minimize this overestimation is to have larger bearing changes in both the X and Y dimension. Such a path is shown in Figure \ref{fig:particletest2} and was used to generate more data for the particle filter to see if the overestimation could be reduced.						

								\begin{figure}[]
						\centering
						\begin{subfigure}[h]{0.95\columnwidth}
							\centering
							\includegraphics[width=.95\columnwidth]{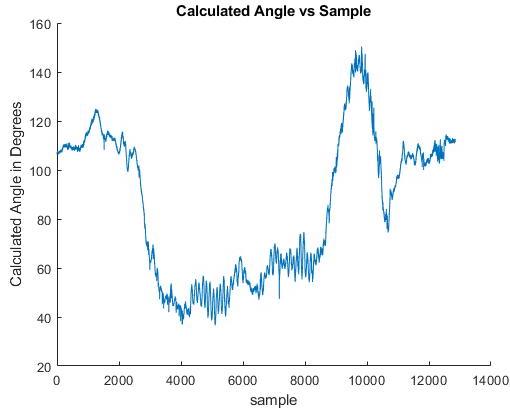}
							\caption{Calculated angle of incidence at full data rate}
							\label{sfig:particletest2movingfulldata}
						\end{subfigure}
						\begin{subfigure}[h]{0.95\columnwidth}
							\centering
							\includegraphics[width=.95\columnwidth]{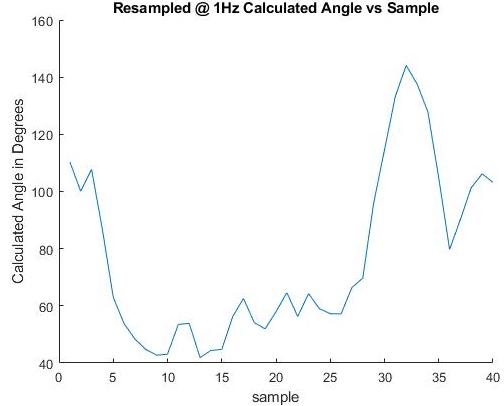}
							\caption{Calculated angle of incidence reduced to 1Hz}
							\label{sfig:particletest2moving1hx}
						\end{subfigure}
					\caption{Moving PDDF array data set circling around signal source}
						\label{fig:main_figuresubsampledparticledata2}
					\end{figure}
					
The second test, shown in Figure \ref{fig:particletest2}, was designed to maximize the bearing change between samples, with the PDDF array moving completely around a stationary signal. 

\begin{figure}[]
						\centering
					
							\centering
							\includegraphics[width=.95\columnwidth]{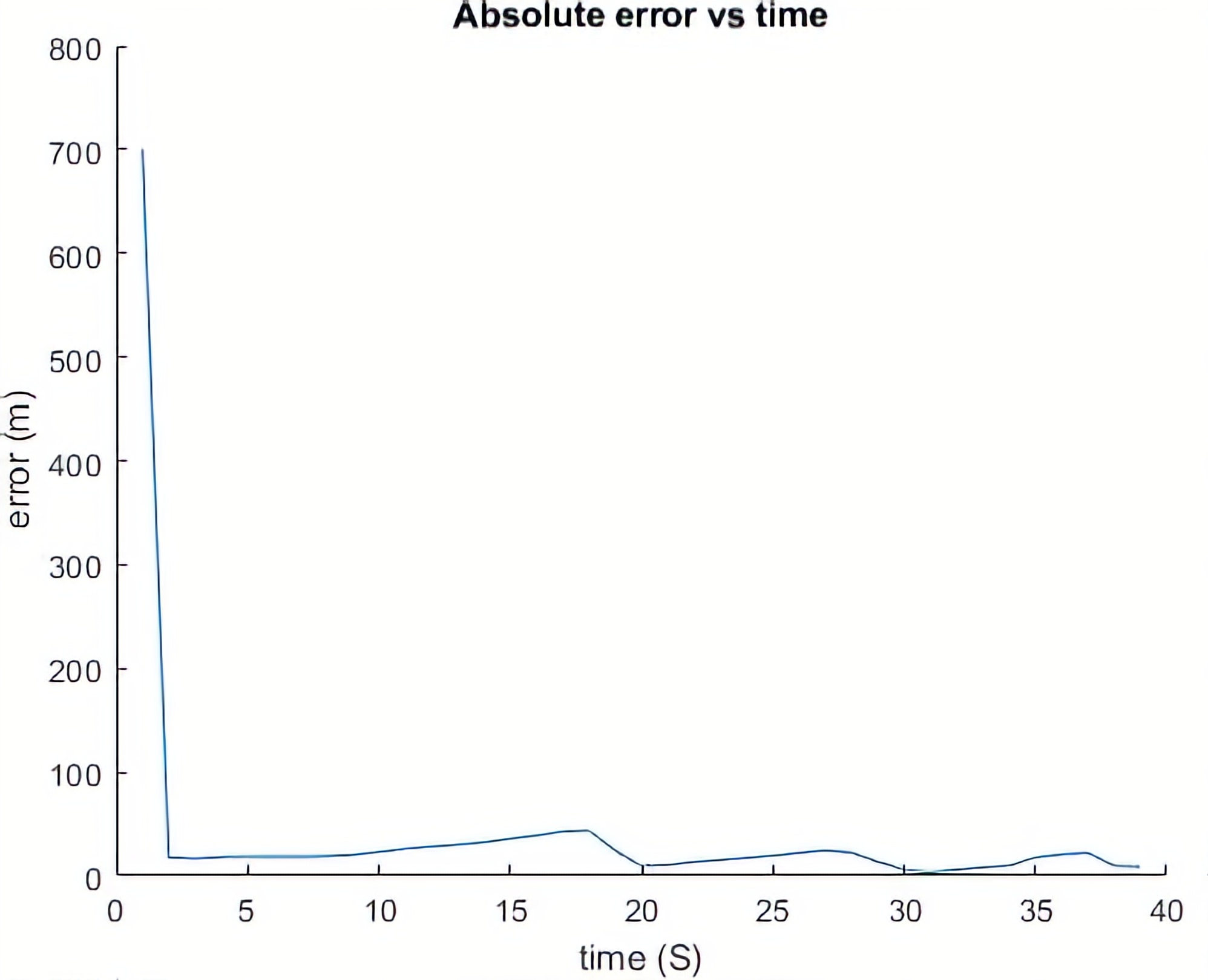}
						
					\caption{Second particle filter results in 2000 meter by 2000 meter grid.}
						\label{fig:main_figureparticlefilter2results2000}
					\end{figure}

The data (Figure~\ref{fig:main_figuresubsampledparticledata2}) was also run through two particle filters as in previous tests. Once again the results were nearly identical and for brevity only the 2km by 2km figures will be shown. The results of the 2km by 2km particle filer are much better and confirms that a rapid change in angle of incidence helps reduce estimated system errors. The total error distance between the actual location and the estimated location was 8.87 meters at the end of the test, and is shown in Figure \ref{fig:main_figureparticlefilter2results2000} and stayed under 40 meters for the duration of the entire test.



Two notes were made about the quality of data collected. Changes in antenna array dimensions and vibrations from the cart introduced ripples in the calculated angle of incidence. In addition, the relative closeness of both tests introduced error from geometric dilution of precision. Longer range testing with a mechanically rigid antenna array could reduce both of these sources of error and increase the position estimate accuracy even more. 

\begin{figure}[]
						\centering
							\centering
										\includegraphics[width=.95\columnwidth]{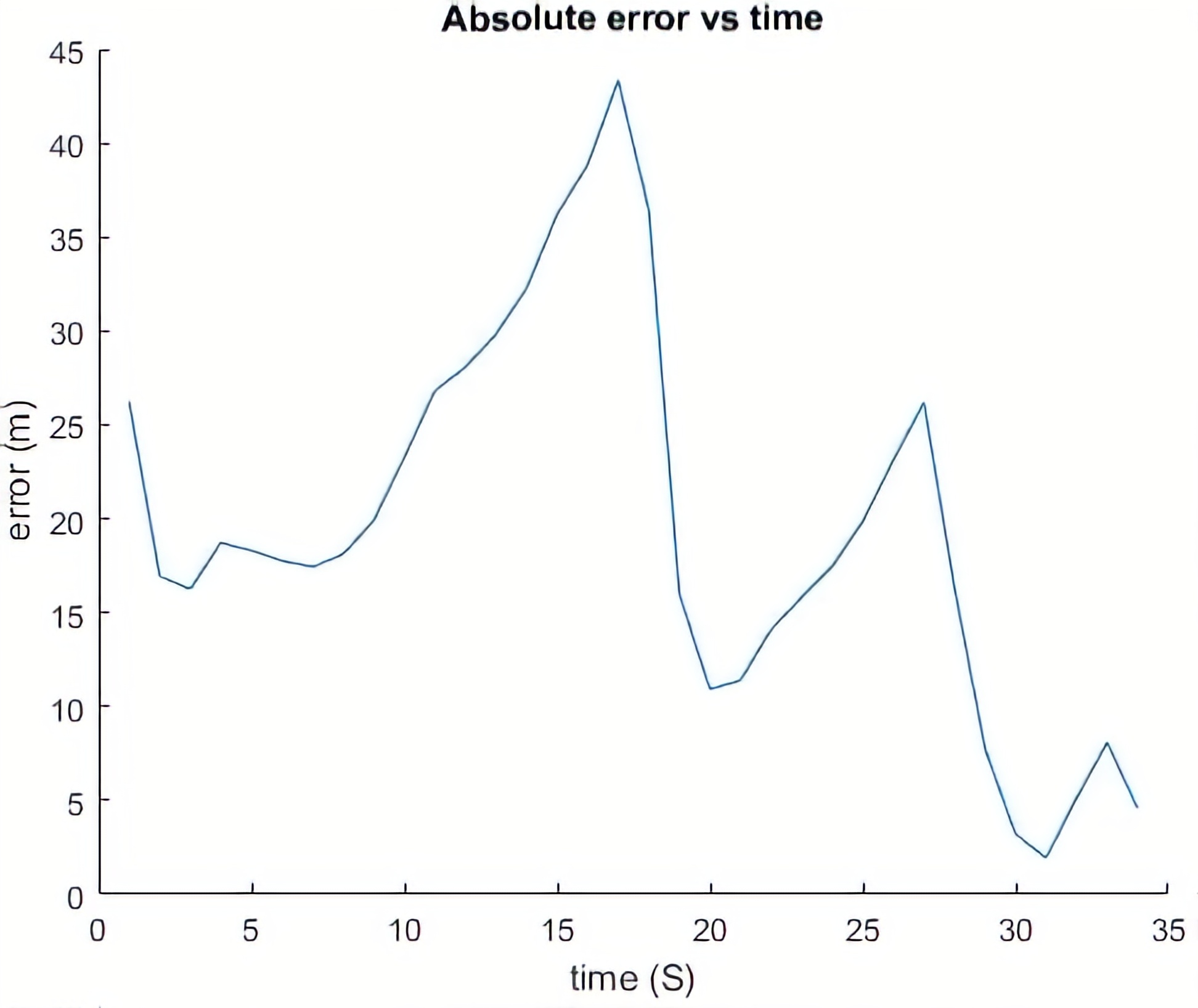}
							
					\caption{Second particle filter tuned results in 200 meter by 200 meter grid.}
						\label{fig:main_figureparticlefilter2results200t}
					\end{figure}

Finally, with some tuning the particle filter's performance, positional estimation error of under 5 meters was achieved as shown in Figure~\ref{fig:main_figureparticlefilter2results200t}, with the data set in Figure \ref{sfig:particletest2moving1hx}. A summary of all the results is found in Table \ref{tab:results}.

			\begin{table}[htb!]
					\centering
					\caption{Particle filter estimation errors at last sample of filter}
					\begin{tabular}{ccccc}
						\toprule
					   Data Set & Grid Size & X Error & Y Error & Total Error \\
						\midrule
						1 & 200 m X 200 m & 21.92 m & 4.45 m & 22.37\\
                       1 & 2 km X 2 km & 21.92 m & 4.52 m & 22.38\\
                        2 & 200 m X 200 m & 8.72 m & 1.47 m & 8.84\\
                        2 & 2 km X 2 km m & 8.58 m & 2.28 m & 8.87\\
                       2 tuned & 200 m X 200 m & 4.57 m & 0.024 m & 4.57\\
                       	\bottomrule
					\end{tabular}
					\label{tab:results}
				\end{table}

\section{Conclusions} \label{ch:conclusions}
	

A low-cost Pseudo Doppler Direction Finder system was presented. This system, unlike other systems in literature, can acquire signals without mechanical rotation and has a 330Hz sample rate. This sample rate is four orders of magnitudes faster than other wildlife tracking techniques~ \cite{torabi_2018, vonehr_2016, bayram_2017,bayram_2018}. When used with the particle filter developed in this work, the PDDF system is capable of tracking a VHF signal's position under 40 meters or error, comparable to other works in this area~\cite{desrochers_2018,bayram_2016}. This high sample rate, when combined with the particle filter and an IMU system, would enable a UAS to track the location of fast-moving wildlife in near real-time. However, two items still need to be completed to enable real-time tracking.

	All systems were functionally, but higher quality IMU data tightly coupled to the bearing measurements would improve the accuracy of the particle filter's localization and enable the use of the PDDF systems full data rate. The current technique requires the data to be down-sampled to match GPS data. Bracing the antennas in a way to prevent changes to the geometry of the array is also necessary to improve angle of incidence accuracy. In addition, a method of rating the quality of the angle of incidence calculation is needed to order to prevent bad data from impacting the particle filter's estimations. Two methods, one uses the HackRF One and one using a different platform, could result in more reliable data acquisition. 

\section{Future Work} \label{s:futurework}
Several avenues for future work exist that would improve the direction finder and the particle filter's performance.
 Foremost, a confidence rating of bearing quality is needed. Even with no incoming signal, or a poor incoming signal not suitable for high accuracy bearing measures, the PDDF system will still generate data. One method to provide that confidence rating using the HackRF One and Operacake combination is measuring the envelope of the output of the notch filter. During testing, it was noted that a good bearing measure always occurred while the output of the notch filter presented a constant envelope. If the output of the notch filter showed fluctuations in amplitude, the bearing result was poor.  By using the envelope’s rate of change as a measure of signal quality, and saving that information with the bearing, the particle filters performance could be improved by increasing the impact of the good bearing data and decreasing the impact of bad bearing data. This confidence rating would also allow multiple non-constant signals to be tracked reliably.

\bibliographystyle{IEEEtran}
\bibliography{IEEEabrv,bibliography}

\begin{thebibliography}{10}
\providecommand{\url}[1]{#1}
\csname url@samestyle\endcsname
\providecommand{\newblock}{\relax}
\providecommand{\bibinfo}[2]{#2}
\providecommand{\BIBentrySTDinterwordspacing}{\spaceskip=0pt\relax}
\providecommand{\BIBentryALTinterwordstretchfactor}{4}
\providecommand{\BIBentryALTinterwordspacing}{\spaceskip=\fontdimen2\font plus
\BIBentryALTinterwordstretchfactor\fontdimen3\font minus
  \fontdimen4\font\relax}
\providecommand{\BIBforeignlanguage}[2]{{%
\expandafter\ifx\csname l@#1\endcsname\relax
\typeout{** WARNING: IEEEtran.bst: No hyphenation pattern has been}%
\typeout{** loaded for the language `#1'. Using the pattern for}%
\typeout{** the default language instead.}%
\else
\language=\csname l@#1\endcsname
\fi
#2}}
\providecommand{\BIBdecl}{\relax}
\BIBdecl

\bibitem{linchant_2015}
J.~Linchant, J.~Lisein, J.~Semeki, P.~Lejeune, and C.~Vermeulen, ``Are unmanned
  aircraft systems (uass) the future of wildlife monitoring? a review of
  accomplishments and challenges,'' \emph{Mammal Review}, vol.~45, no.~4, p.
  239–252, 2015.

\bibitem{vonehr_2016}
K.~Vonehr, S.~Hilaski, B.~E. Dunne, and J.~Ward, ``Software defined radio for
  direction-finding in uav wildlife tracking,'' \emph{2016 IEEE International
  Conference on Electro Information Technology (EIT)}, p. 0464–0469, May
  2016.

\bibitem{bayram_2017}
H.~Bayram, N.~Stefas, K.~S. Engin, and V.~Isler, ``Tracking wildlife with
  multiple uavs: System design, safety and field experiments,'' \emph{2017
  International Symposium on Multi-Robot and Multi-Agent Systems (MRS)}, p.
  97–103, Dec 2017.

\bibitem{bayram_2018}
H.~Bayram, N.~Stefas, and V.~Isler, ``Aerial radio-based telemetry for tracking
  wildlife,'' \emph{2018 IEEE/RSJ International Conference on Intelligent
  Robots and Systems (IROS)}, p. 4723–4728, Oct 2018.

\bibitem{isler2015finding}
V.~Isler, N.~Noori, P.~Plonski, A.~Renzaglia, P.~Tokekar, and J.~Vander~Hook,
  ``Finding and tracking targets in the wild: Algorithms and field
  deployments,'' in \emph{2015 IEEE International Symposium on Safety,
  Security, and Rescue Robotics (SSRR)}.\hskip 1em plus 0.5em minus 0.4em\relax
  IEEE, 2015, pp. 1--8.

\bibitem{vander2012cautious}
J.~Vander~Hook, P.~Tokekar, and V.~Isler, ``Cautious greedy strategy for
  bearing-based active localization: Experiments and theoretical analysis,'' in
  \emph{2012 IEEE International Conference on Robotics and Automation}.\hskip
  1em plus 0.5em minus 0.4em\relax IEEE, 2012, pp. 1787--1792.

\bibitem{tokekar2013tracking}
P.~Tokekar, E.~Branson, J.~Vander~Hook, and V.~Isler, ``Tracking aquatic
  invaders: Autonomous robots for monitoring invasive fish,'' \emph{IEEE
  Robotics \& Automation Magazine}, vol.~20, no.~3, pp. 33--41, 2013.

\bibitem{bayram_2016}
H.~Bayram, K.~Doddapaneni, N.~Stefas, and V.~Isler, ``Active localization of
  vhf collared animals with aerial robots,'' \emph{2016 IEEE International
  Conference on Automation Science and Engineering (CASE)}, p. 934–939, Aug
  2016.

\bibitem{desrochers_2018}
A.~Desrochers, J.~Tremblay, Y.~Aubry, D.~Chabot, P.~Pace, and D.~Bird,
  ``Estimating wildlife tag location errors from a vhf receiver mounted on a
  drone,'' \emph{Drones}, vol.~2, no.~4, p. 44–53, Dec 2018.

\bibitem{maccurdy2011automated}
R.~B. MacCurdy, R.~M. Gabrielson, and K.~A. Cortopassi, ``Automated wildlife
  radio tracking,'' \emph{SA Zekavat \& RM Buehler (red). Handbook of position
  location: theory, practice, and advances. Wiley, Hoboken, NJ}, 2011.

\bibitem{dopplerdf}
``Doppler df instruments : More picodopp df information,''
  \url{http://www.silcom.com/~pelican2/PicoDopp/PICO_MORE.html}, (Accessed on
  02/29/2020).

\bibitem{Rohde_Schwarz}
``Introduction into theory of direction finding,'' \emph{RohdeI \& Schwarz
  Radiomonitoring \& Radiolocation | Catalog 2016}, p. 62–85, 2016.

\bibitem{HackRFOn58:online}
``Hackrf one - great scott gadgets,''
  \url{https://greatscottgadgets.com/hackrf/one/}, (Accessed on 09/15/2019).

\bibitem{OperaCak4:online}
``Opera cake · mossmann/hackrf wiki,''
  \url{https://github.com/mossmann/hackrf/wiki/Opera-Cake}, (Accessed on
  09/15/2019).

\bibitem{moell_2008}
\BIBentryALTinterwordspacing
J.~Moell, ``Wide-range antenna arrays for the roanoke doppler,'' Apr 2008.
  [Online]. Available: \url{http://www.homingin.com/newdopant.html}
\BIBentrySTDinterwordspacing

\bibitem{maarell2002foraging}
A.~M{\aa}rell, J.~P. Ball, and A.~Hofgaard, ``Foraging and movement paths of
  female reindeer: insights from fractal analysis, correlated random walks, and
  l{\'e}vy flights,'' \emph{Canadian Journal of Zoology}, vol.~80, no.~5, pp.
  854--865, 2002.

\bibitem{benhamou2007many}
S.~Benhamou, ``How many animals really do the l{\'e}vy walk?'' \emph{Ecology},
  vol.~88, no.~8, pp. 1962--1969, 2007.

\bibitem{bartumeus2005animal}
F.~Bartumeus, M.~G.~E. da~Luz, G.~M. Viswanathan, and J.~Catalan, ``Animal
  search strategies: a quantitative random-walk analysis,'' \emph{Ecology},
  vol.~86, no.~11, pp. 3078--3087, 2005.

\bibitem{johansen_2009}
A.~M. Johansen, ``Smctc: Sequential monte carlo in c++,'' \emph{Journal of
  Statistical Software}, vol.~30, no.~6, 2009.

\bibitem{chen_1998}
J.~S. Liu and R.~Chen, ``Sequential monte carlo methods for dynamic systems,''
  \emph{Journal of the American Statistical Association}, vol.~93, no. 443, p.
  1032–1044, 1998.

\bibitem{torabi_2018}
A.~Torabi, M.~W. Shafer, G.~S. Vega, and K.~M. Rothfus, ``Uav-rt: An sdr based
  aerial platform for wildlife tracking,'' \emph{2018 IEEE 88th Vehicular
  Technology Conference (VTC-Fall)}, Aug 2018.

\end{thebibliography}

\end{document}